\documentclass[conference]{IEEEtran}
\IEEEoverridecommandlockouts
\usepackage{cite}
\usepackage{amssymb,amsfonts}
\usepackage{indentfirst} 
\usepackage{amsmath} 
\usepackage{cases}
\usepackage{graphicx}
\usepackage{textcomp}
\usepackage{xcolor}
\usepackage{amsmath} 
 
\newtheorem{remark}{Remark} 
\newtheorem{proposition}{Proposition}

\usepackage{subfigure} 
\usepackage{algorithm}
\usepackage{algorithmic}
\def\BibTeX{{\rm B\kern-.05em{\sc i\kern-.025em b}\kern-.08em
    T\kern-.1667em\lower.7ex\hbox{E}\kern-.125emX}}
\begin{document}

\title{An Efficient Method for Sample Adversarial Perturbations against Nonlinear Support Vector Machines\\

}

\author{\IEEEauthorblockN{ Wen Su}
\IEEEauthorblockA{\textit{School of Mathematics and Statistics} \\
\textit{Beijing Institute of Technology}\\
Beijing, China \\
suwen019@163.com}
%\and
%\IEEEauthorblockN{2\textsuperscript{nd} Chunfeng Cui}
%\IEEEauthorblockA{\textit{LMIB of the Ministry of Education} \\
%\textit{School of Mathematical Sciences} \\
%\textit{Beihang University}\\
%Beijing, 100191, China \\
%chunfengcui@buaa.edu.cn}
\and \IEEEauthorblockN{ Qingna Li\textsuperscript{*}\thanks{\textsuperscript{*}Corresponding author. This author's research is supported by the National Natural Science Foundation of China (NSFC) 12071032.
}} \IEEEauthorblockA{\textit{School of Mathematics and Statistics/} \\ \textit{Beijing Key Laboratory on MCAACI/}\\ \textit{Key Laboratory of Mathematical Theory }\\ \textit{and Computation in Information Security}\\	 \textit{Beijing Institute of Technology}\\ Beijing,  China \\ qnl@bit.edu.cn}
}
		
\maketitle

\begin{abstract}
Adversarial perturbations have drawn great attentions in various machine learning models. In this paper, we investigate the sample adversarial perturbations for nonlinear support vector machines (SVMs). Due to the implicit form of the nonlinear functions mapping data to the feature space, it is difficult to obtain the explicit form of the adversarial perturbations. By exploring the special property of nonlinear SVMs, we transform the optimization  problem of attacking nonlinear SVMs into a nonlinear KKT system. Such a system can be solved by various numerical methods.  Numerical results show that our method is efficient in computing adversarial perturbations.
%Adversarial perturbations have drawn great attentions in various machine learning models. Most of them are calculated by iteration, which takes a long time and can not be well explained. In this paper, we investigate the adversarial perturbations for support vector machines(SVMs). Through the Random Fourier Features method, we transform the optimization  problem of attacking nonlinear SVM into the optimization problem of attacking linear SVM in feature space, so as to obtain the exact directions of perturbations in feature space.
%Our method avoids a large number of iterations and possible non convergence in the iterative process. Numerical results show that our method is fast and effective in computing adversarial perturbations.
\end{abstract}

\begin{IEEEkeywords}
adversarial perturbation, support vector machine,  KKT system, nonlinear optimization, machine learning
\end{IEEEkeywords}

\section{Introduction}
Machine learning tools are often applied to real-world problems such as computer vision \cite{ref101,ref102}, natural language processing \cite{ref2,ref103,ref104}, recommender systems \cite{ref105,ref106}, internet of things\cite{ref107,ref108}, etc., and have achieved great success. However, existing research has found that machine learning models, even the state-of-the-art deep neural networks, are vulnerable to attack. By analyzing the vulnerabilities of machine learning models, attackers add small perturbations to the input data, and it can lead the machine learning system to making wrong predictions about the data with high probability \cite{b0}. This stimulates the enthusiasm for research on adversarial perturbations. Adversarial perturbations can be classified into three types: perturbations for a given sample (sAP), universal adversarial perturbations (uAP) and class-universal adversarial perturbations (cuAP). Below we briefly review them one by one.

sAP is one of the most significant and frequently-used types of adversarial perturbations. Since 2014, researchers have begun to study sAP in deep neural networks, and proposed many methods to calculate sAP. Szegedy et al. firstly discovered a surprising weakness of neural networks, neural networks are vulnerable to very small adversarial perturbations \cite{b1}. In the observation of human vision system, the attacked image with these small perturbations is almost indistinguishable from the original clean image, which is the original sAP.
Goodfellow et al. proposed the Fast Gradient Sign Method (FGSM) \cite{b2}. Since that, sAP, which is constructed according to the gradient information of the model, has become one of the popular methods of perturbation \cite{b3,b4}. Deepfool is an attack method, based on linearization and separating hyperplane \cite{b5}. 
  In more extreme cases,  \cite{b5-1} propose a novel method for generating one-pixel adversarial perturbations, which requires less information.  
uAP is another popular type of adversarial perturbations. 
Moosavi-Dezfooli et al. first proposed a single small perturbation, which can fool state-of-the-art deep neural network classifiers on all natural images \cite{b6}. This perturbation is called uAP, and it is added to the entire dataset, rather than a single sample. Since \cite{b6}, researchers have proposed different methods to generate uAP, including data-driven and data-independent methods \cite{b8,b9,b1-2}. 
As for cuAP, it is an modified uAP based on different applications.  It can secretly attack the data in a class-discriminative way. In \cite{b1-1}, the authors noticed that uAP might have obvious impacts, which makes users suspicious. Thus, they proposed  cuAP, which is referred to as class discriminative universal adversarial perturbation therein.

Recently, several studies have focused on adversarial perturbations added to SVMs.  \cite{b10,b11} obtain the approximate solution of the perturbations on various SVMs, and give the robustness analysis of SVMs through experiments.
Su et al. \cite{b2-2} focused on considering the special structure of linear SVMs, and gave several types of analytical solutions to  perturbations against SVMs. However, for nonlinear SVMs, how to develop adversarial perturbations is not addressed.

In this paper, we mainly study adversarial perturbations on nonlinear SVMs. Due to the implicit form of the nonlinear functions mapping data to the feature space, it is difficult to obtain explicit form of the adversarial perturbations. By explaining the special property of nonlinear SVMs, we transform the optimization  problem of attacking the nonlinear SVM into a nonlinear KKT system. Such a system can be solved by various numerical methods.  Numerical results show that our method is efficient in computing adversarial perturbations. 

The organization of the paper is as follows. In Section \ref{Optimization Model for Sample Adversarial Perturbations against Binary Nonlinear SVMs}, we  introduce the optimization models of sAP for binary nonlinear SVMs. In Section \ref{Feasibility Study}, we discuss the theoretical properties of the optimization model as well as how to solve it. Finally, we present some numerical result in Section \ref{Numerical Experiments} and draw  conclusions in Section \ref{conclusion}.

\section{Optimization Model for Sample Adversarial Perturbations against Binary Nonlinear SVMs}\label{Optimization Model for Sample Adversarial Perturbations against Binary Nonlinear SVMs}

\subsection{The Setting of Nonlinear SVMs.}

For the binary classification problem, we assume that the training data set is $ T=\{(x_1,y_1),$  $ (x_2,y_2), \dots,$   $(x_m,y_m)\} $, with $ x_i \in \mathbb{R}^d $ and $ y_i \in \{-1,1\} $, $ i=1, 2, \dots, m $ is the label of the corresponding $ x_i $. Let $ \phi :\mathbb{R}^d   \rightarrow \mathbb{X}$ be the nonlinear mapping which maps the data in $ \mathbb{R}^d$ to the feature space $ \mathbb{X} $.
The decision function of nonlinear SVM trained on dataset $ T $ is given by 
\begin{equation}
k(x)=sign(w^T \phi (x)+b).
\end{equation}
Here $ sign(t) $ is defined by the sign of $ t $, which is $ 1 $ if $ t>0 $, $ -1 $ if $ t<0 $.
Usually, $ k $ is trained via the dual approach, and $ (w, b) $ takes the following form
\begin{equation}
w=\sum_{i=1}^{m}\lambda_i y_i \phi (x_i), \ 
\end{equation}
\begin{equation}
b=\dfrac{1}{|S|}\sum_{s \in S} \left(y_s-\sum\limits_{j \in S} \lambda_j y_j \left\langle \phi (x_s), \phi (x_j)\right\rangle  \right),
\end{equation}
where $ \lambda \in \mathbb{R}^m $ is the solution  of the corresponding dual problem of nonlinear SVMs,    $ S=\{i \ | \ \lambda_i>0, i=1, 2, \dots, m\}$ is the index set of all support vectors.
This gives the following decision function:
\begin{equation}\label{nonlinear1} 
\begin{split}
{k}(x)=&sign\left(\sum_{i=1}^{m}\lambda_i y_i \left\langle \phi (x_i), \phi (x)\right\rangle +b \right) \\
=&sign\left(\sum_{i=1}^{m}\lambda_i y_i K(x, x_i) +b\right), 
\end{split}
\end{equation} 
where the kernel function $ K: \mathbb{R}^d \times \mathbb{R}^d \rightarrow \mathbb{R} $ is defined as $ K(x, x') = \left\langle\phi (x), \phi (x')\right\rangle$. 
Popular kernel functions include:
\begin{itemize} 	
	\item Linear kernel: $ K(x, x') = x^T x'$,
	\item Polynomial kernel: $ K(x, x') = (p x^T x'+a)^b$, $ p>0$, 	\item Radial basis function (RBF) kernel: $ K(x, x') = exp(-\gamma \|x-x'\|^2)$, $ \gamma > 0$. 
\end{itemize}

\subsection{Optimization model for sAP.}
 Below we discuss the situation of sAP. That is, assume that the nonlinear SVM is trained on the data set $ T $, with the resulting classifier given in \eqref{nonlinear1}. For a given data $ \hat{x} \in \mathbb{R}^d$, the  optimization model for generating the adversarial perturbation against   the trained SVMs  is  to look for a perturbation  $ r \in \mathbb{R}^d $ with the smallest length $ \|r\|_2 $, such that the data $ \hat{x} \in  \mathbb{R}^d $ can be misclassified \cite{b2-2}. That is, the label after perturbed by $ r $  (that is $ {k}(\hat{x}+r) $)  is different from the unperturbed label $  {k}(\hat{x}) $. It leads to the following optimization model:
\begin{equation}\label{equ2}
	\begin{aligned}
	\min_{r \in \mathbb{R}^d} \  & \  \|r\|_2  \\
	\hbox{s.t.} \  & \  {k}(\hat{x}+r)\neq {k}(\hat{x}), 
	\end{aligned}
\end{equation}
which is equivalent to the following problem
\begin{equation}\label{nonlinear3}
	\begin{aligned}
	\min_{r \in \mathbb{R}^d} \  & \  \|r\|_2^2  \\
	\hbox{s.t.} \  & \  {k}(\hat{x}+r)\neq {k}(\hat{x}). 
	\end{aligned}
\end{equation}
Consider the case that $ {k}(\hat{x})=1 $. That is, 
\begin{equation}\label{new3}
\sum_{i=1}^{m}\lambda_i y_i K(\hat{x}, x_i) +b>0.
\end{equation}
Then  \eqref{nonlinear3} reduces to the following form
\begin{equation}\label{nonlinear4} 	
\begin{aligned} 	
\min_{r \in \mathbb{R}^d} \  & \  \|r\|_2^2  \\ 	\hbox{s.t.} \  & \  {k}(\hat{x}+r)=-1. 	
\end{aligned} 
\end{equation} 
\eqref{nonlinear4} is equivalent to the following problem
\begin{equation}\label{nonlinear100} 	
\begin{aligned} 	
\min_{r \in \mathbb{R}^d} \  & \  \|r\|_2^2  \\ 	\hbox{s.t.} \  & \  \sum_{i=1}^{m}\lambda_i y_i K(\hat{x}+r, x_i) +b <0. 	
\end{aligned} 
\end{equation}

%If $ {k}(\hat{x}+r)=0 $, we regard that $ \hat{x}+r $ is not misclassified. That is, we require $ {k}(\hat{x}+r)=-1 $, which is equivalent to $ \sum_{i=1}^{m}\lambda_i y_i K(\hat{x}+r, x_i) +b<0 $.
 To make sure that $ \hat{x}+r $ is classified wrongly, it is natural to solve the following problem (given a small scale $ \varepsilon >0 $)
\begin{equation}\label{nonlinear5} 	
	\begin{aligned} 	
	\min_{r \in \mathbb{R}^d} \  & \  \|r\|_2^2  \\ 	\hbox{s.t.} \  & \  \sum_{i=1}^{m}\lambda_i y_i K(\hat{x}+r, x_i) +b \leq -\varepsilon,	
	\end{aligned} 
\end{equation} 
where we actually require that $ \hat{x} $ is surely misclassified in the sense that $  {k}(\hat{x}+r)=-1  $.
Obviously, $ r=0 $ is not a feasible solution, since $ \sum_{i=1}^{m}\lambda_i y_i K(\hat{x}, x_i) +b+ \varepsilon >  \varepsilon > 0$ due to \eqref{new3}.

Next, we will derive the KKT conditions for \eqref{nonlinear5}. The Lagrange function of  \eqref{nonlinear5} is as follows
\begin{equation}\label{nonlinear6}  
L(r,\mu)= \|r\|_2^2+\mu \left(\sum_{i=1}^{m}\lambda_i y_i K(\hat{x}+r, x_i) +b+ \varepsilon\right),
\end{equation}
where $ \mu \in \mathbb{R} $ is the   Lagrange multiplier corresponding to the inequality constraint in \eqref{nonlinear5}.

The KKT conditions of \eqref{nonlinear5} are given below
\begin{subequations} 	\begin{numcases}{}  	\nabla_r L(r,\mu)=2r+\mu \nabla_r\left(\sum_{i=1}^{m}\lambda_i y_i K(\hat{x}+r, x_i)\right)=0 \label{nonlinear7_1} \\ \mu \left(\sum_{i=1}^{m}\lambda_i y_i K(\hat{x}+r, x_i) +b+ \varepsilon \right)=0  \label{nonlinear7_2} \\ \mu \geq 0, \ \sum_{i=1}^{m}\lambda_i y_i K(\hat{x}+r, x_i) +b+ \varepsilon \leq 0. \label{nonlinear7_3}	\end{numcases} 	\label{nonlinear7} \end{subequations} 

Next, we discuss some preliminary facts about the KKT system \eqref{nonlinear7}.
Notice that if $ \mu = 0 $, by \eqref{nonlinear7_1}, there is $ r=0 $. In this case, as we mentioned above, there is 
\begin{equation}
\!\sum_{i=1}^{m}\lambda_i y_i K(\hat{x}+r, x_i) \!
	= \! \sum_{i=1}^{m}\lambda_i y_i K(\hat{x}, x_i) +b+ \varepsilon 
	\! > \! \varepsilon \! > \! 0,
\end{equation}
 which contradicts with \eqref{nonlinear7_3}. Therefore, $ \mu \neq 0 $.

Consequently, by  \eqref{nonlinear7_2}, we have 
\begin{equation}\label{nonlinear8}   \sum_{i=1}^{m}\lambda_i y_i K(\hat{x}+r, x_i) +b+ \varepsilon=0. \end{equation}
In other words, the KKT conditions reduce to the following statement. We are looking for  $ (r, \mu) \in \mathbb{R}^{d+1}$, such that the following system holds
\begin{equation}\label{nonlinear9} 	 
\begin{cases} 	      
2r+\mu \nabla_r\left(\sum\limits_{i=1}^{m}\lambda_i y_i K(\hat{x}+r, x_i)\right)=0  \\ 	\sum\limits_{i=1}^{m}\lambda_i y_i K(\hat{x}+r, x_i) +b+ \varepsilon=0 \\
\mu > 0.	 		   
\end{cases} 
\end{equation}

\section{Theoretical Properties of \eqref{nonlinear5}.}\label{Feasibility Study}
\subsection{Feasibility Study.}
Now our aim is to solve the optimization problem \eqref{nonlinear5}, which reduces to \eqref{nonlinear9}. Before we discuss how to solve \eqref{nonlinear9}, we need to stop for a while to take a look at the feasibility issue of \eqref{nonlinear5}. That is, whether the feasible set of problem \eqref{nonlinear5} is empty. This issue is important due to the following reason. If there is no feasible point $ r $ that satisfies the constraint in \eqref{nonlinear5}, then it is not meaningful to solve the corresponding equivalent system \eqref{nonlinear9}. Below, we present our result in Proposition \ref{proposition1}.

\begin{proposition}\label{proposition1}
	Let $ k(x) $ be trained on dataset $ T $ by the nonlinear SVM with kernel function $ K $.
	(\romannumeral1) The feasible set of problem \eqref{nonlinear5} is not empty. (\romannumeral2) Moreover, let $ (\overline{r}, \overline{\mu}) $ be one solution of the KKT system \eqref{nonlinear7}. Then the linearly independent constraint qualification (LICQ) holds at $ \overline{r} $ .
\end{proposition}
Proof. 
To show the feasibility of \eqref{nonlinear5}, we need to find a vector $ r \in \mathbb{R}^{d}$, which satisfies the inequality in \eqref{nonlinear5}. Define the following two sets
\begin{equation}
I_{+}=\left\{ i \in \{1, 2, \dots, m\} \ | \ y_i=1, \ {k}(x_i)>0\right\},
\end{equation}
\begin{equation}
I_{-}=\{ i \in \{1, 2, \dots, m\} \ | \ y_i=-1, \ {k}(x_i)<0\}.
\end{equation}
In other words, the data $ x_i, i \in I_{+} $ and the data $ x_i, i \in I_{-} $ are correctly classified by $ k(x) $, since their labels are the same as their predicted labels. A basic fact is that $ I_{+} \neq \emptyset $ and $ I_{-} \neq \emptyset $.
Define $ \overline{c}$ and $ \varepsilon $ as follows:
\begin{equation}\label{new1}
	\overline{c} = \max\limits_{i \in I_{-}} {k}(x_i), \  \varepsilon=  -\frac{1}{2}\overline{c}.
\end{equation}
 By the definition of $ I_{-} $, there is $ \overline{c} <0 $ and consequently $ \varepsilon>0 $. Furthermore, for all $ i \in I_{-}$, there is ${k}(x_i) \leq \max\limits_{i \in I_{-}} {k}(x_i)=\overline{c}$. It follows that 
 \begin{equation}
 	{k}(x_i)+ \varepsilon \leq \overline{c}+\varepsilon= \frac{1}{2}\overline{c} <0, \ i \in I_{-}.
 \end{equation} 
 Since $ I_{-} \neq \emptyset $, we can pick up an index $ i_0 \in  I_{-}$. Let $ r_0=x_{i_0}- \hat{x} \in \mathbb{R}^{d}$. 
 One can verify that 
 \begin{equation}\label{nonlinear10}  
 \begin{split} {k}(\hat{x}+r_0)+ \varepsilon=&{k}(x_{i_0}- \hat{x} + \hat{x})+ \varepsilon\\
 =&{k}(x_{i_0})+\varepsilon\\
  \leq&\overline{c}-\frac{1}{2}\overline{c}\\
 =&\frac{1}{2}\overline{c}\\
 <&0. \end{split} 
 \end{equation}
 In other words, we found a feasible solution of problem \eqref{nonlinear5}, which satisfies the inequality in \eqref{nonlinear5} strictly. Therefore, the feasible set of problem \eqref{nonlinear5} is not empty.
 
%Due to the above statement, we know that the Slater’s condition holds for the contraint in \eqref{nonlinear5}. 
To solve \eqref{nonlinear9}, let $ (\overline{r}, \overline{\mu}) $ be the solution of the KKT system. By our preliminary analysis, we know that there is $ \sum_{i=1}^{m}\lambda_i y_i K(\hat{x}+\overline{r}, x_i)+b+ \varepsilon=0$ and $\mu>0 $. Also we know that $ \overline{r} \neq 0 $. Next, we only need to show that 
\begin{equation}\label{new2}
\nabla_r\left(\sum_{i=1}^{m}\lambda_i y_i K(\hat{x}+\overline{r}, x_i)\right) \neq 0.
\end{equation}

It indeed holds by noting \eqref{nonlinear7_1}. Since $ 2\overline{r}=- \mu \nabla_r\left(\sum_{i=1}^{m}\lambda_i y_i K(\hat{x}+\overline{r}, x_i)\right)$.
Together with the fact that $ \overline{r} \neq 0 $ and $ \overline{\mu}>0 $, we get \eqref{new2}. Therefore,  LICQ holds for each $ \overline{\mu} $ which satisfies the KKT condition \eqref{nonlinear7}. The proof is finished.
\hfill$\Box$

\begin{remark}
	Note that problem \eqref{nonlinear5} is in general a nonlinear optimization program with one nonlinear inequality constraint. Therefore, it is possible that problem \eqref{nonlinear5} is nonconvex. The result in Proposition \ref{proposition1} guarantees that the optimal solution of problem \eqref{nonlinear5} is one of the KKT points satisfying \eqref{nonlinear7}. Below, we discuss how to solve the equivalent system \eqref{nonlinear9} numerically.
\end{remark}

\subsection{Numerical algorithms for \eqref{nonlinear9}.}
To solve the KKT system
\eqref{nonlinear9}, define
\begin{equation}\label{nonlinear11} 	  
F(r,\mu)=\begin{bmatrix} 	       2r+\mu \sum\limits_{i=1}^{m}\lambda_i y_i \nabla_r K(\hat{x}+r, x_i)\\ 	\sum\limits_{i=1}^{m}\lambda_i y_i K(\hat{x}+r, x_i) +b+ \varepsilon \end{bmatrix}.
\end{equation}
\eqref{nonlinear9} reduces to looking for $ (r, \mu) $, such that 
\begin{equation}\label{new16}
	F(\mu)=0 \ \mbox{and} \ \mu>0. 
\end{equation}

To deal with the inequality constraint $ \mu >0 $ in \eqref{new16}, we can further reformulate \eqref{new16} as the following system 
\begin{equation}\label{new17} 	   H(r,\mu)=\begin{bmatrix} 	       2r+ |\mu| \sum\limits_{i=1}^{m}\lambda_i y_i \nabla_r K(\hat{x}+r, x_i)\\ 	\sum\limits_{i=1}^{m}\lambda_i y_i K(\hat{x}+r, x_i) +b+ \varepsilon \end{bmatrix}.
\end{equation}
By replacing $ \mu $ by $ |\mu| $ in the first equation of \eqref{new17}, one can see that we only need to solve \eqref{new17} instead. We have the following result addressing the relationship of the solutions for \eqref{new16} and \eqref{new17}.

\begin{proposition}\label{proposition2} 	Let $ (r^{\ast}, {\mu}^{\ast}) $ be the solution of nonlinear system  \eqref{new17}.
Then $ (r^{\ast}, |{\mu}^{\ast}|) $ is the solution of  \eqref{new16}.
\end{proposition}

Proof. 
It is trivial.
\hfill$\Box$ 

%If K is the RBF kernel function, there is $$ \nabla_r K(\hat{x}+r, x_i)= -2 \gamma  (\hat{x}+r-x_i) exp(-\gamma \|\hat{x}+r-x_i\|^2), \ \gamma>0. $$
%
%
%If K is Polynomial kernel function, there is $$ \nabla_r K(x+r, x_i)= bp x_i (p (\hat{x}+r)^T x_i+a)^{b-1}, \ p>0.$$

%We reach the following algorithm for \eqref{nonlinear9}.
%我们得到了 \eqref{nonlinear9} 的以下算法。
{We get the following algorithm to describe the complete process of  generating adversarial perturbation.}
\begin{algorithm}[H]
	\caption{The complete process of  generating adversarial perturbation} %算法的名字
	{\bf Input:} %算法的输入， \hspace*{0.02in}用来控制位置，同时利用 \\ 进行换行
	The training data set $ (x_i,y_i) $,  $ x_i \in \mathbb{R}^d $, $ i=1, 2, \dots, m $, the kernel function K, the fixed small scale $ \varepsilon >0 $. Given data $ \hat{x} \in \mathbb{R}^d $.\\
	{\bf Output:} %算法的结果输出
	Adversarial perturbation $ r $.
	
		\textbf{Step 1.} Train the dataset to obtain the SVM model $ {k}(x)=sign\left(\sum_{i=1}^{m}\lambda_i y_i K(x, x_i) +b\right)  $. That is, obtain $ \lambda_i$, $ i=1, 2, \dots, m $ and $ b $.
		
		\textbf{Step 2.}  Replace the parameters in \eqref{new17} with the trained $ \lambda_i$, $ i=1, 2, \dots, m $, $ b $ and the kernel function K. Then solve the  nonlinear equations system \eqref{new17} to get $ (r^*,\mu^*) $, i.e., $ r_0 $.  Then $ (r^*, |\mu^*|) $ is the KKT point satisfying \eqref{new16}.
		
		\textbf{Step 3.} Output $ r^* $.
\end{algorithm}

\subsection{Extension to the case of $ {k}(\hat{x})=-1 $.}
Note that all our above discussions work in the case of $ {k}(\hat{x})=1 $. We can extend it similarly to the case of $ {k}(\hat{x})=-1 $. We briefly describe it below.

If $ {k}(\hat{x})=-1 $,  problem \eqref{nonlinear5} is replaced by the following problem
\begin{equation}\label{new5} 	
\begin{aligned} 	
\min_{r \in \mathbb{R}^d} \  & \  \|r\|_2^2  \\ 	\hbox{s.t.} \  & \  -\sum_{i=1}^{m}\lambda_i y_i K(\hat{x}+r, x_i) -b \leq -\varepsilon,	
\end{aligned} 
\end{equation}
and the  KKT conditions of \eqref{new5} are given below
\begin{equation}\label{new8} 	  
\begin{cases} 
 2r-\mu \nabla_r\left(\sum\limits_{i=1}^{m}\lambda_i y_i K(\hat{x}+r, x_i)\right)=0 \\ 
\mu \left(-\sum\limits_{i=1}^{m}\lambda_i y_i K(\hat{x}+r, x_i) -b+ \varepsilon \right)=0\\
\mu \geq 0, \ -\sum\limits_{i=1}^{m}\lambda_i y_i K(\hat{x}+r, x_i) -b+ \varepsilon \leq 0.
\end{cases}  
\end{equation}
Similarly, \eqref{new8} can be reduced to
\begin{equation}\label{new9} 	 
\begin{cases} 	      
2r-\mu \nabla_r\left(\sum\limits_{i=1}^{m}\lambda_i y_i K(\hat{x}+r, x_i)\right)=0  \\ 	-\sum\limits_{i=1}^{m}\lambda_i y_i K(\hat{x}+r, x_i) -b+ \varepsilon=0 \\
\mu > 0.	 		   
\end{cases} 
\end{equation}
We only need to solve the following nonlinear system
\begin{equation}\label{new11} 	  
\begin{bmatrix} 	       2r- |\mu| \sum\limits_{i=1}^{m}\lambda_i y_i \nabla_r K(\hat{x}+r, x_i)\\ 	-\sum\limits_{i=1}^{m}\lambda_i y_i K(\hat{x}+r, x_i) -b+ \varepsilon \end{bmatrix}=0.
\end{equation}
In a word, for $ {k}(\hat{x})=-1 $, we only need to replace \eqref{new17} by \eqref{new11} in Algorithm 1.

 Combining the two cases, we use pseudocode in Algorithm \ref{alg1} to explain the working principle of Algorithm 1.

\begin{algorithm}
	%\textsl{}\setstretch{1.8}
	\renewcommand{\algorithmicrequire}{\textbf{Input:}}
	\renewcommand{\algorithmicensure}{\textbf{Output:}}
	\caption{Pseudocode of  adversarial perturbation design for nonlinear SVMs}
	\label{alg1}
	\begin{algorithmic}[1]
		\REQUIRE Image $x$, classifier $ k $, $ \varepsilon $.
		\ENSURE Adversarial perturbation $ r $.
		\STATE Initialize $\hat{x} \gets x$. 
		\IF{${k}(\hat{x})=1$} 
		\STATE $(r,\mu)  \!  \gets \! $ Solve  $\begin{bmatrix} 	       2r+ |\mu| \sum\limits_{i=1}^{m}\lambda_i y_i \nabla_r K(\hat{x}+r, x_i)\\ 	\sum\limits_{i=1}^{m}\lambda_i y_i K(\hat{x}+r, x_i) +b+ \varepsilon \end{bmatrix} \! =0 $. 
		\ELSE 
		\STATE $(r,\mu)  \!  \gets   \!  $ Solve $ \begin{bmatrix} 	       2r- |\mu| \sum\limits_{i=1}^{m}\lambda_i y_i \nabla_r K(\hat{x}+r, x_i)\\ 	-\sum\limits_{i=1}^{m}\lambda_i y_i K(\hat{x}+r, x_i) -b+ \varepsilon \end{bmatrix} \! =0 $. 		
		\ENDIF  		
		\RETURN r.
	\end{algorithmic}  
\end{algorithm}

\section{Numerical Experiments}\label{Numerical Experiments}

In this section, we conduct extensive numerical test to verify the efficiency of our method.  First, we will introduce the two datasets we used in the experiment. 
All experiments are tested in Matlab R2019b in Windows 10 on a HP probook440 G2 with an Intel(R) Core(TM) i5-5200U CPU at 2.20 GHz and of 4 GB RAM.
All classifiers are trained using the LIBSVM\footnote{It can be downloaded from https://www.csie.ntu.edu.tw/$ \sim $cjlin/libsvm.}, where the $ L_1\mbox{-}$loss kernel SVM model is used. 

For the nonlinear system \eqref{nonlinear11}, we use the \texttt{fslove} package in MATLAB to solve \eqref{nonlinear11} with command $ [x,fval,exitflag]=fsolve(@H(r, \mu),x0) $.  It basically solve \eqref{new17} by choosing the methods between `trust-region-dogleg' (default), `trust-region', and `levenberg-marquardt'. Here, we use the option 'trust-region-dogleg'.
 
\subsection{Experimental data settings}

MNIST was first applied in \cite{b3-2}, and CIFAR-10 was first applied in \cite{b3-3}. They are the most popularly used in the experiment of relevant papers related to adversarial perturbation. Recent papers on adversarial perturbation applying MNIST are \cite{b1,b2,b3,b4,b5,b10,b11,b2-2}, and the papers  apply CIFAR-10 are\cite{b2,b3,b4,b5,b5-1,b1-1,b10,b2-2}. Therefore, we test the adversarial perturbations against SVMs on MNIST and CIFAR-10 image classification datasets. The specific explanations of these two datasets are as follows.
\begin{itemize}
	\item MNIST: The complete MNIST dataset has a total of 60,000 training samples and 10,000 test samples, each of which is a vector of 784 pixel values and can be restored to a $ 28*28 $ pixel gray-scale handwritten digital picture. The value of the recovered handwritten digital picture ranges from 0 to 9, which exactly corresponds to the 10 labels of the dataset. 
	\item CIFAR-10: CIFAR-10 is a color image dataset closer to universal objects. The complete CIFAR-10 dataset has a total of 50,000 training samples an 10,000 test samples, each of which is a vector of 3072 pixel values and can be restored to a $ 32* 32 *3 $ pixel RGB color picture. There are 10 categories of pictures, each with 6000 images. The picture categories are airplane, automobile, bird, cat, deer, dog, frog, horse, ship and truck, their labels correspond to $ \{ 0, 1, 2, 3, 4, 5, 6, 7, 8, 9 \} $ respectively.
\end{itemize}

\subsection{Role of Parameters}

\subsubsection{$ \varepsilon $ in \eqref{new17}}

\begin{figure}[h]  \centerline{\includegraphics[width=0.5\textwidth]{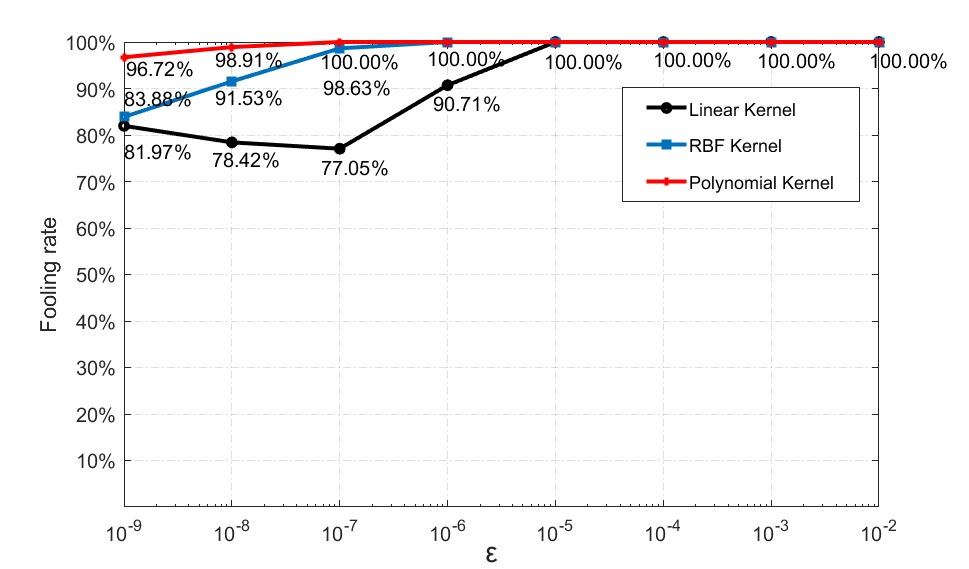}}  \caption{The  fooling rate for  kernel SVMs with variation of $ \varepsilon $.}  \label{fig0}  \end{figure}

In this experiment, the data is selected from MNIST dataset with the positive class 1  and the negative class 0. We use LIBSVM to establish linear, RBF  and polynomial binary classification SVMs for the above data. We  calculate the fooling rate of adversarial perturbations  for data with the change of $ \varepsilon $. The fooling rate is defined as the percentage of samples whose prediction changes after adversarial perturbations is applied, i.e., 
\begin{equation}
	\dfrac{\text{The number of } {k}( \hat{x}+r)\neq {k}(\hat{x}), \hat{x} \in A}{\text{The number of data in } A },
\end{equation}
where $A$ is a given dataset and $ {k} $ is the trained SVM classifier.
We show the results in Fig.~\ref{fig0}. Here $ A $ is chosen as a  dataset with 366 data.

We found that in the case of linear kernel, when $ \varepsilon $ increases, the fooling rate will decrease first, and increase when $ \varepsilon $ is greater than $ 10^{-7} $. When $ \varepsilon $ is greater than $ 10^{-5} $, the fooling rate can reach $ 100\% $.
However, in the case of RBF kernel and polynomial kernel, when $ \varepsilon $ increases, the fooling rate increases. The difference is in the case of RBF kernel, when $ \varepsilon $ is greater than $ 10^{-6} $, the fooling rate can reach $ 100\% $, but in the case of polynomial kernel, when $ \varepsilon $ is greater than $ 10^{-7} $, the fooling rate can reach $ 100\% $.
%Therefore, in order to ensure that the perturbation can successfully attack all data, and the choice of $ \varepsilon $ will not be too large, in the following experiments, we choose $ \varepsilon = 10^{-5} $ in the case of linear kernel, $ \varepsilon = 10^{-6} $ in the case of RBF kernel, $ \varepsilon = 10^{-7} $ in the case of polynomial kernel.
Therefore, considering the above three cases comprehensively, in order to ensure that the perturbation can successfully attack all data, and the choice of $ \varepsilon $ will not be too large, in the following experiments, we choose $ \varepsilon = 10^{-5} $.

%We found that in the case of linear kernel, when $ \varepsilon $ increases, the fooling rate will decrease first, and increase when $ \varepsilon $ is greater than $ 10^{-7} $. When $ \varepsilon $ is greater than $ 10^{-5} $, the fooling rate can reach $ 100\% $.
%Therefore, in order to ensure that the perturbation can successfully attack all data, and the choice of $ \varepsilon $ will not be too large, in the following experiments, we choose $ \varepsilon = 10^{-5} $.
%
%We get that in the case of RBF kernel, when $ \varepsilon $ increases, the fooling rate increases. When $ \varepsilon $ is greater than $ 10^{-6} $, the fooling rate can reach $ 100\% $. Therefore, in order to ensure that the adversarial perturbations can successfully attack all data and the choice of $ \varepsilon $ will not be too large, we choose $ \varepsilon = 10^{-6} $ in this part of the experiment.
%
%We find that in the case of polynomial kernel, when $ \varepsilon $ increases, the fooling rate will gradually increase. When $ \varepsilon $ is greater than $ 10^{-7} $, the fooling rate can reach $ 100\% $. In this part of the experiment, we choose $ \varepsilon = 10^{-7} $.

\subsubsection{$ \gamma $ in RBF Kernel}

\begin{figure}[htbp]  	\centerline{\includegraphics[width=0.35\textwidth]{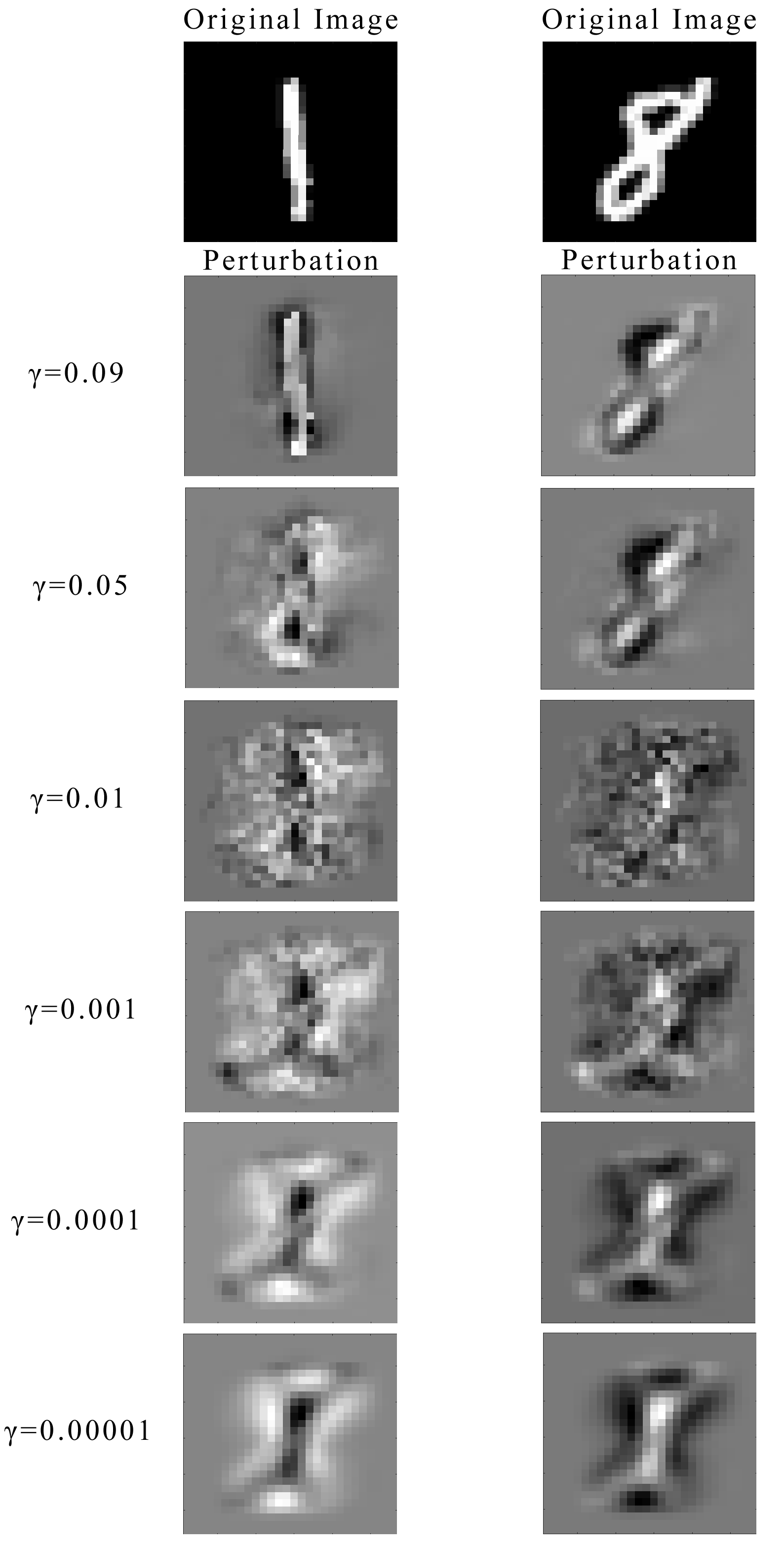}}  	\caption{Comparison of adversarial perturbations between positive and negative classes when $ \gamma $ changes.}  	\label{fig4}  \end{figure}

%The data of this experiment were selected from MNIST dataset, the positive class selects the number 1 in the dataset, and the negative class selects the number 8, a total of 12593. We use LIBSVM to establish RBF kernel nonlinear binary SVMs  for the above data.
To see the role of $ \gamma$ in RBF kernel, we conduct some further test  with dataset number 1 and number 8.  $ \gamma $ is chosen from the set $\Gamma= \{0.00001, 0.0001, 0.001, 0.01, 0.05, 0.09\} $, then we can get six classifiers. We select a fixed data of number 1 and a fixed data of number 8 as the original clean data of positive and negative classes respectively, and we show them at the top of  Fig.~\ref{fig4}. Then we solve the  nonlinear equations system \eqref{new17}, and calculate  sAP corresponding to the SVM model when $ \gamma \in \Gamma$, and show them in  Fig.~\ref{fig4}.

As shown in  Fig.~\ref{fig4}, after adding the pertubations, the corresponding images' classes become 8 and 1. We found that even if the  range of  $ \gamma $ is large, our methods have solutions, which means we can get the corresponding attacks, to make the classifier be fooled successfully. At the same time, with the increase of  $ \gamma $, the obtained sAP gradually approaches the shape of the number itself.
When the  $ \gamma $ is the same, comparing the positive and negative classes horizontally, we find that the adversarial perturbations generated by the positive and negative classes have more properties of themselves and less properties of the other.

\subsubsection{$ b $ and $ p $ in Polynomial Kernel}

\begin{figure}[htbp]  	\centerline{\includegraphics[width=0.39\textwidth]{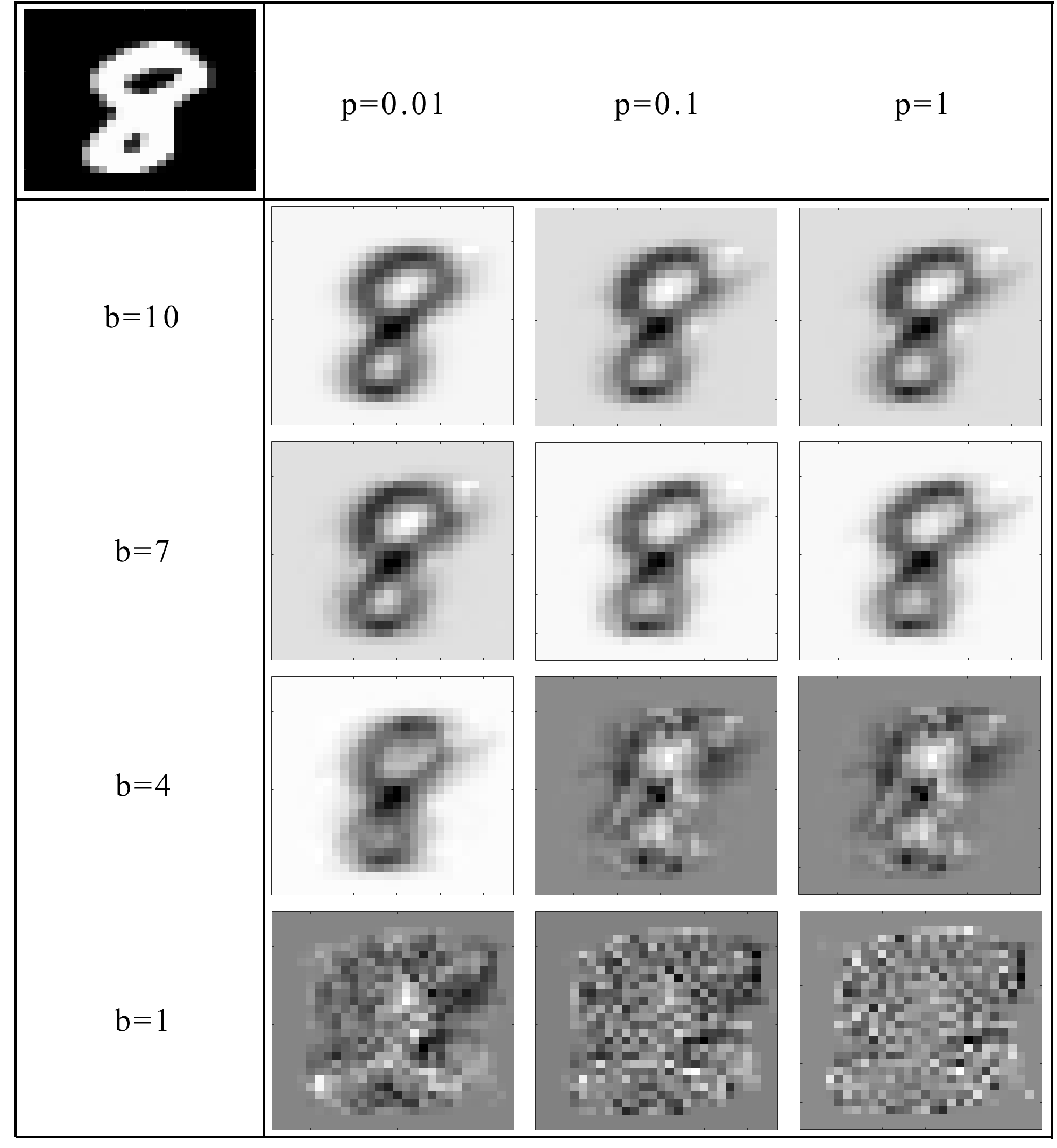}}  	\caption{For different $ b $ and $ p $, the change of adversarial perturbations of the same number 8.}  	\label{fig6}  \end{figure}

To see the role of parameters $ b $ and $ p $ in the polynomial kernel, we conduct further test with dataset number 1 and number 8.
We select a fixed number 8 as the original clean image and put it in the upper left of Fig.~\ref{fig6}. As we all know, $ b $ and $ p $ of polynomial kernel determine its nonlinearity. 
As the default setting in LIBSVM, we fix parameter $ a $ as 0. Then we select $ b $ from $ \{1,4,7,10\} $, and select $ p $ from $ \{0.01,0.1,1\} $. The adversarial perturbations corresponding to different combinations of $ b $ and $ p $ are demonstrated in  Fig.~\ref{fig6}.

As shown in  Fig.~\ref{fig6}, when $ p $ is fixed, the adversarial perturbations is closer to the original image with the increase of $ b $. When $ b $ is fixed, with the decrease of $ p $, the adversarial perturbations is closer to the original image. It can be seen from Fig.~\ref{fig6} that both $ b $ and $ p $ have an impact on the adversarial perturbations, and $ b $ has a greater impact.

\begin{figure}[htbp]  	\centerline{\includegraphics[width=0.39\textwidth]{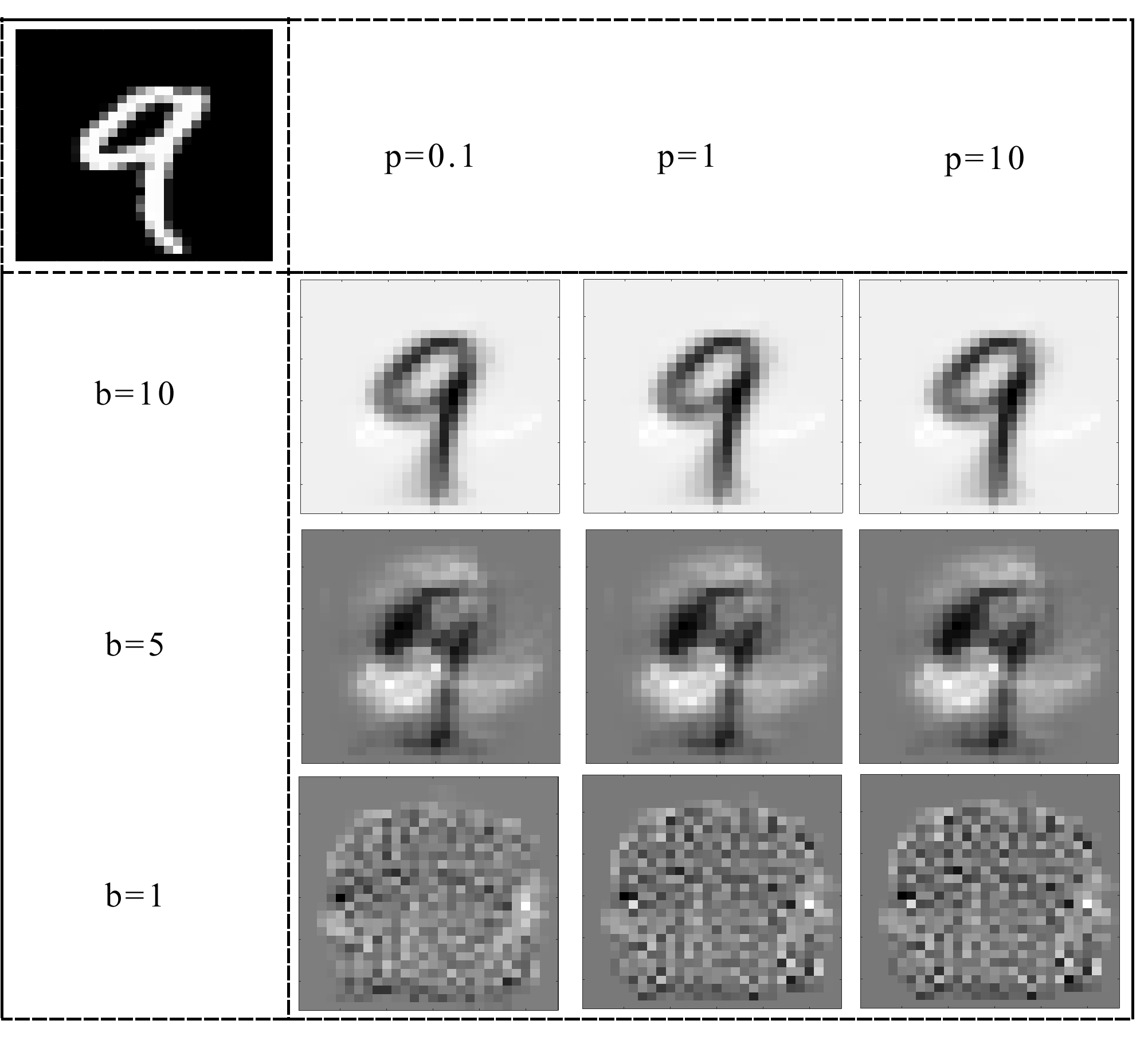}}  	\caption{For different $ b $ and $ p $, the change of adversarial perturbations of the same number 9.}  	\label{fig7}  \end{figure}

A similar test is conducted with dataset number 2 and number 9, with $ a=0 $,  $b \in \{1,5,10 \} $ and $p \in \{0.1,1,10\}$. The result is reported in Fig.~\ref{fig7}. It can be seen that when
$p$ is fixed, the adversarial perturbations is closer to the original image with the increase of $b$. When $b$ is fixed, the change of $p$ has little effect on the adversarial perturbations.

\subsection{Comparison of Methods for linear kernel SVM}

\begin{figure}[bhtp]
	\centering
	\subfigure[Results by EC method.]
	{
		\begin{minipage}[b]{.46\linewidth}
			\centering
			\includegraphics[scale=0.16]{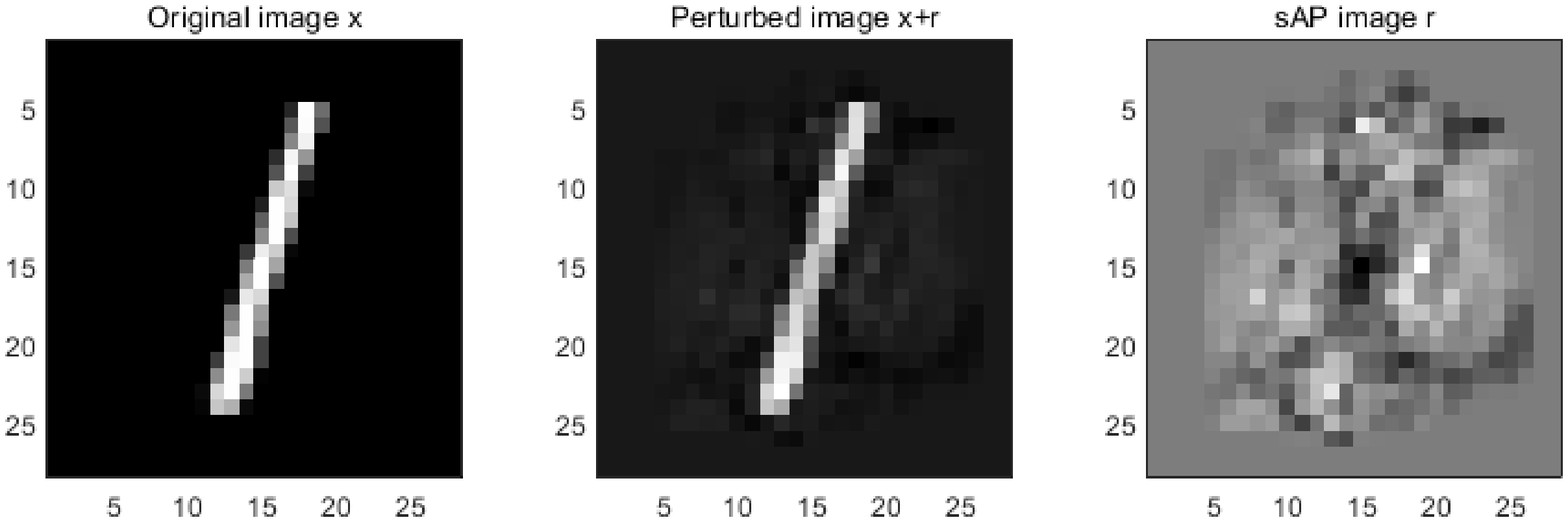} \\
			\includegraphics[scale=0.16]{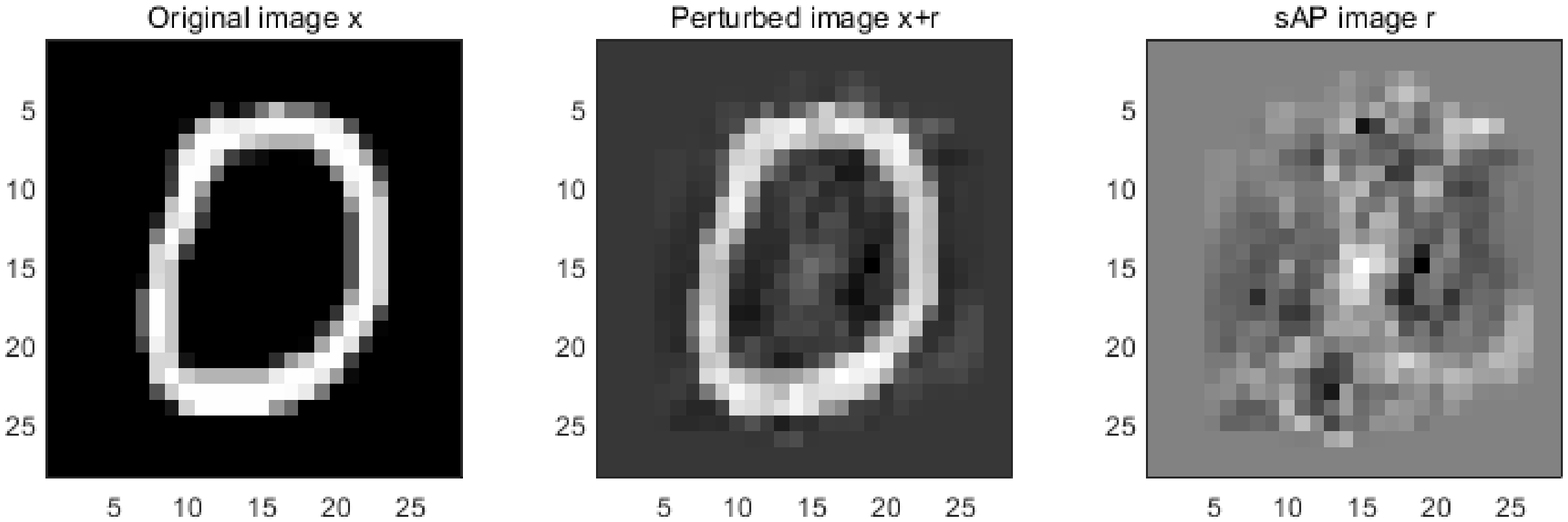}
		\end{minipage}
	}
	\subfigure[Results by Algorithm \ref{alg1}.]
	{
		\begin{minipage}[b]{.46\linewidth}
			\centering
			\includegraphics[scale=0.16]{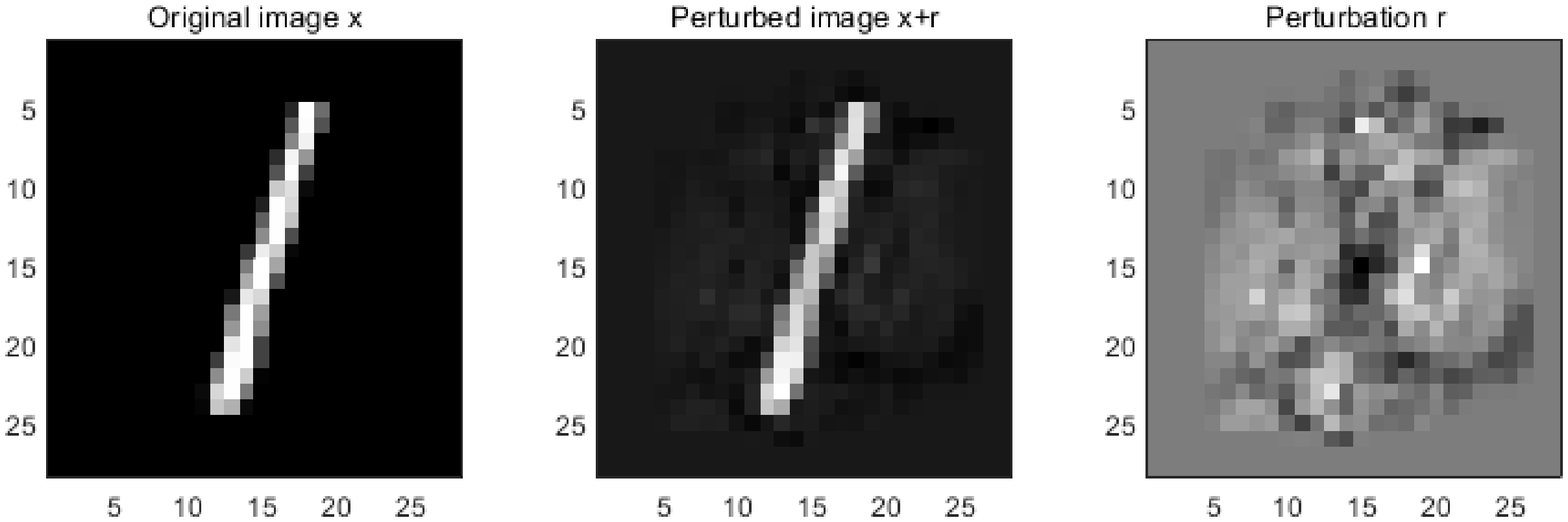} \\
			\includegraphics[scale=0.16]{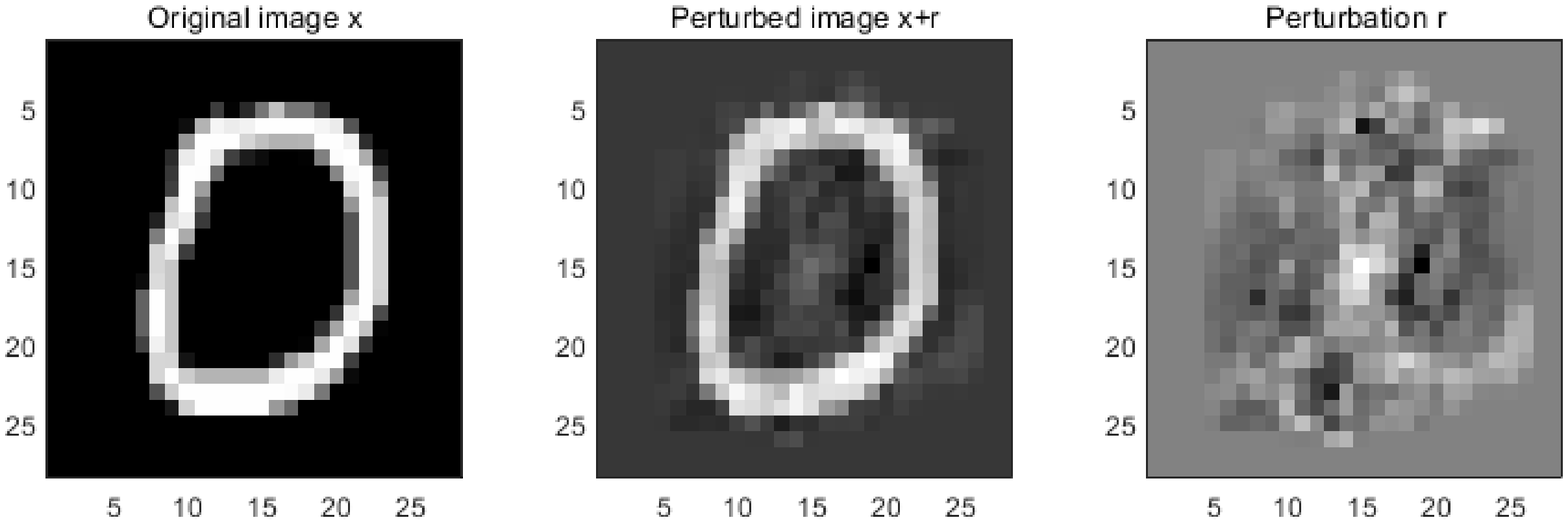}
		\end{minipage}
	}
\caption{The original image, the image that has been misclassified after being attacked, and the  image of sAP against linear kernel SVM.}
	\label{fig2}
\end{figure}

\begin{figure}[bhtp]
	\centering
	\subfigure[Results by EC method.]
	{
		\begin{minipage}[b]{.46\linewidth}
			\centering
			\includegraphics[scale=0.16]{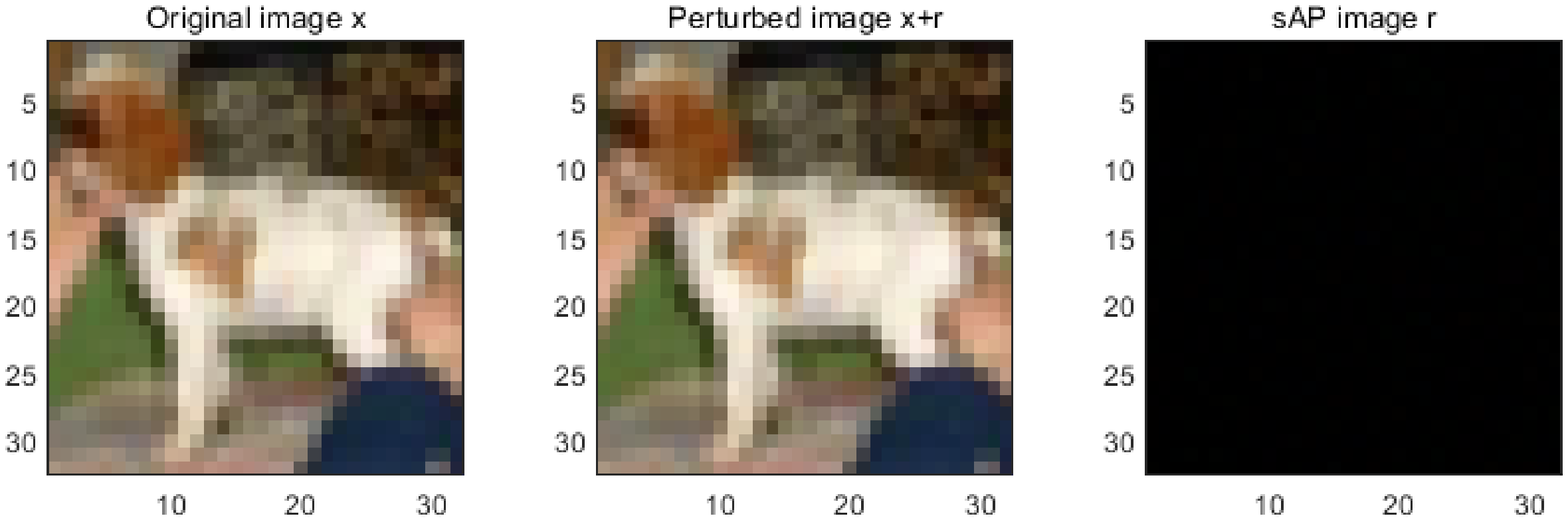} \\
			\includegraphics[scale=0.16]{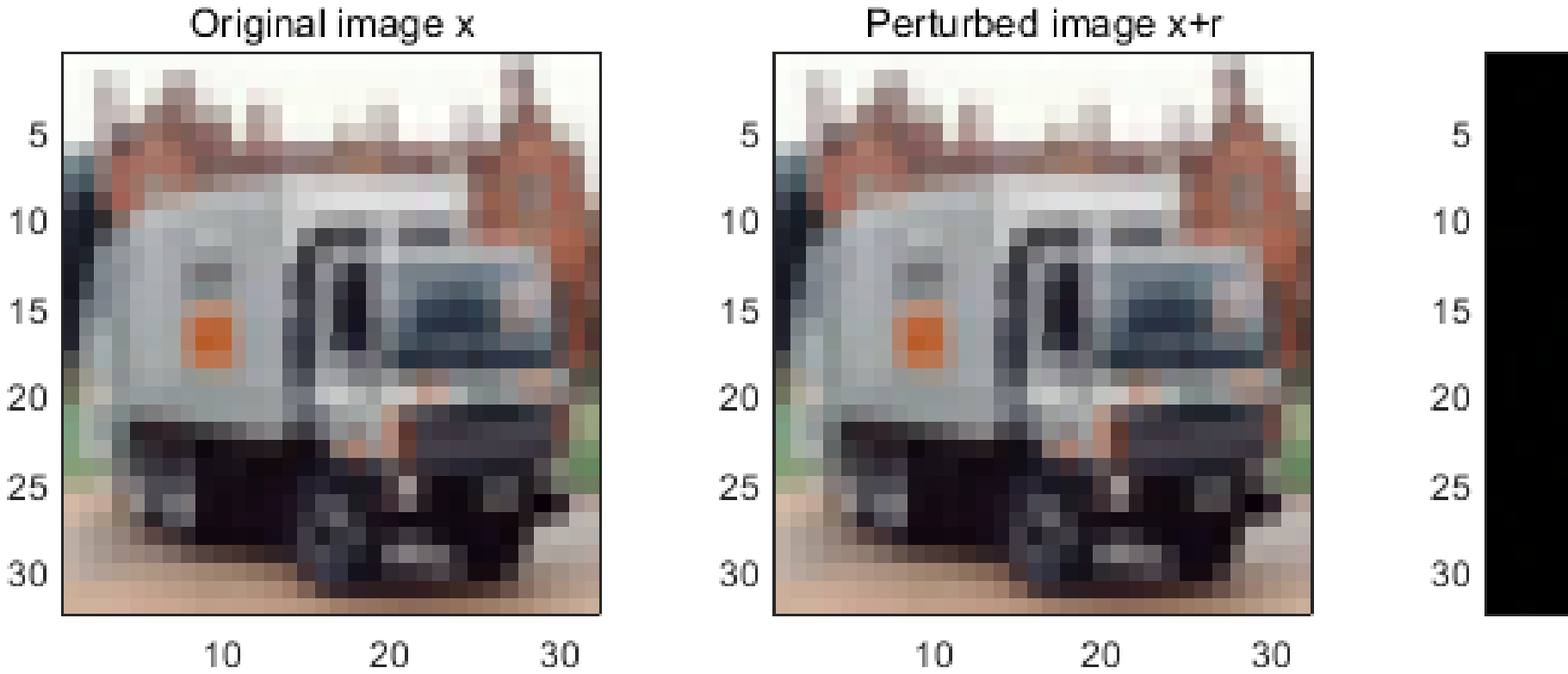}
		\end{minipage}
	}
	\subfigure[Results by Algorithm \ref{alg1}.]
	{
		\begin{minipage}[b]{.46\linewidth}
			\centering
			\includegraphics[scale=0.16]{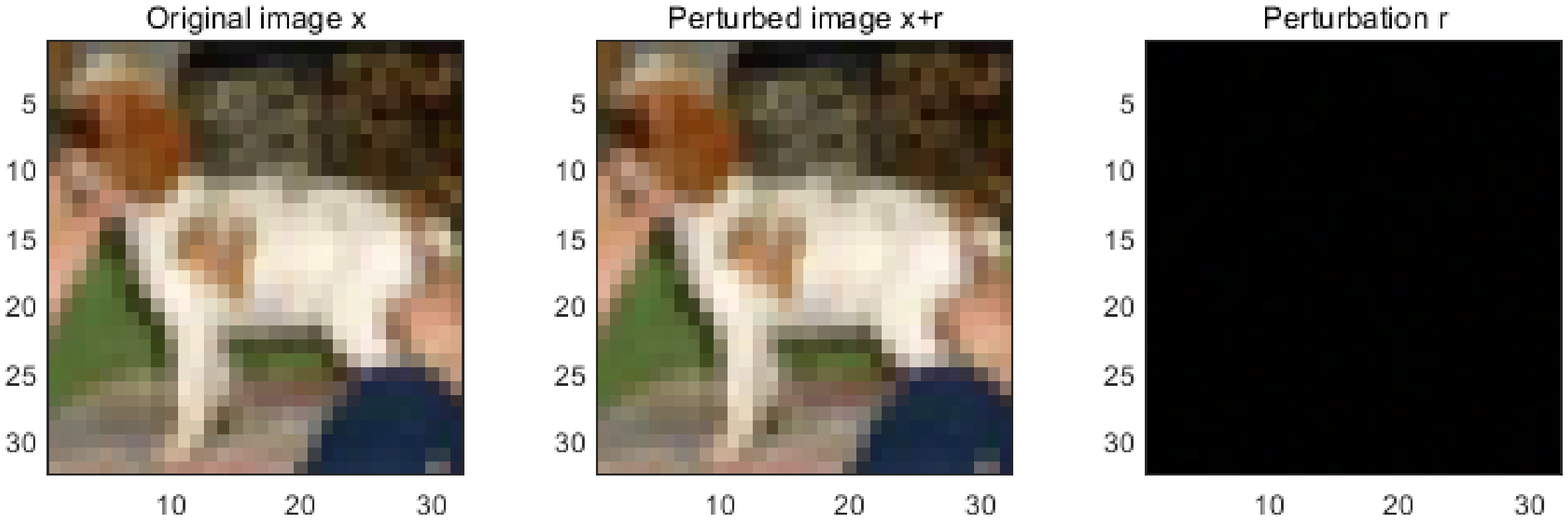} \\
			\includegraphics[scale=0.16]{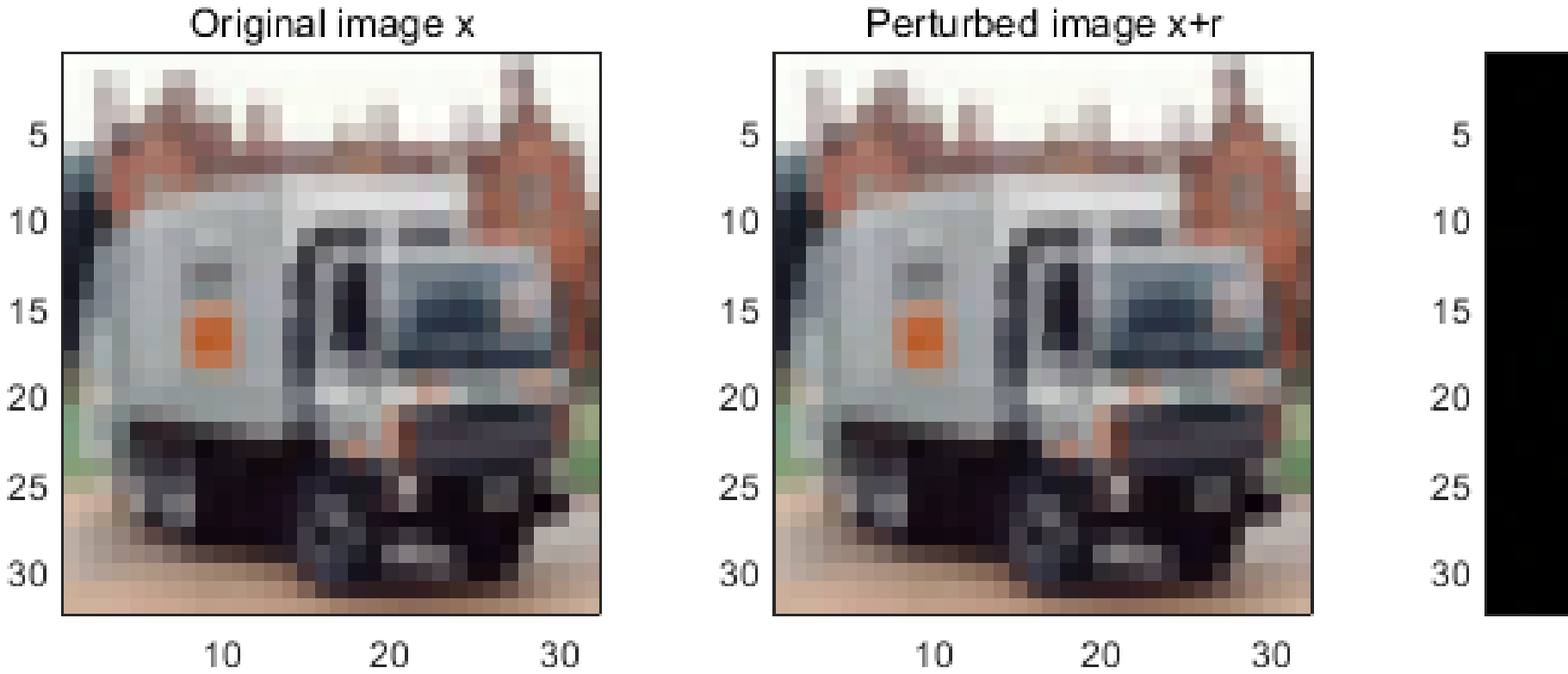}
		\end{minipage}
	}
	\caption{The original image, the image that has been misclassified after being attacked, and the  image of sAP against  linear kernel SVM.}
	\label{fig2new}
\end{figure}

In \cite{b2-2}, we give the explicit formulae of sAP, uAP and cuAP for linear SVM.  In Fig.~\ref{fig2} and Fig.~\ref{fig2new}, we use the explicit calculation in \cite{b2-2} (short for EC) and Algorithm \ref{alg1} to solve sAP respectively. 
In Fig.~\ref{fig2}, we give two examples to compare the original image,  the image misclassified after being attacked, and the image of sAP of MNIST dataset.   At the top of Fig.~\ref{fig2}, the original number of the image is 1. After adding the attacks obtained by two methods respectively, the number of the perturbed image becomes 0. 
%The calculation CPU time of the EC method is $ 0.032s $, and the calculation CPU time of Algorithm 1  is $ 4.335s $.
%It is reasonable since we need to solve a nonlinear system by an iterative solver in Algorithm 1, where as the EC method provide explicit calculation of sAP.    
Obviously, the attacks obtained by the two methods are the same, and in human eyes, the class of the perturbed images does not change.
At the bottom of  Fig.~\ref{fig2}, the original number of the image is 0. After adding the attacks obtained by the two methods respectively, the number of the perturbed image becomes 1. 
%and the calculation CPU time  of Algorithm 1 is $ 3.399s $.

Similarly, at the top of Fig.~\ref{fig2new}, the original image is dog. After adding the attack, the  perturbed image becomes truck. At the bottom of  Fig.~\ref{fig2new},  the original image is truck and  the  perturbed image is dog.   
%The average calculation CPU time of the EC method is $ 0.926s $, and the average calculation CPU time of Algorithm 1  is $ 1.351 \times 10^{3}s $.

\subsection{Numerical experiments of CIFAR-10 }

\begin{figure}[h] 
	\centerline{\includegraphics[width=0.35\textwidth]{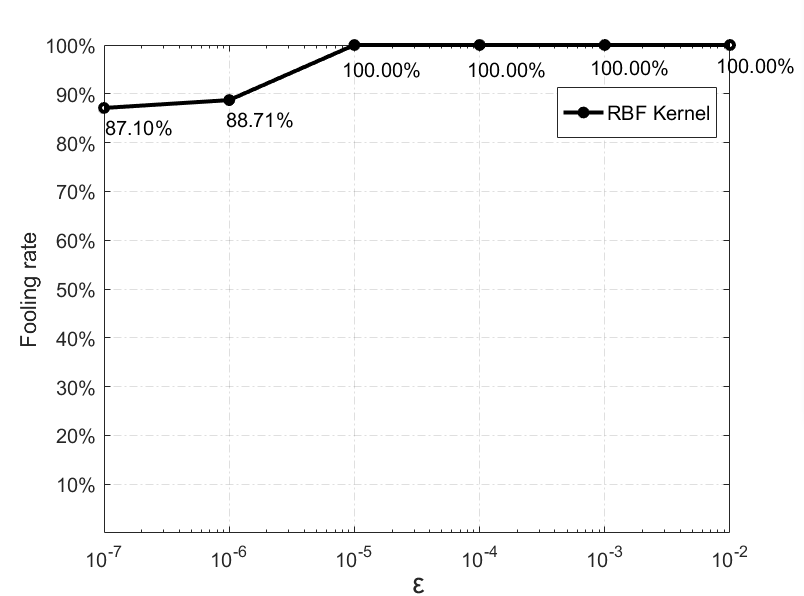}} 
	\caption{The  fooling rate for RBF kernel SVM with variation of $ \varepsilon $.} 
	\label{fig8} 
\end{figure}

For CIFAR-10,  we conduct experiment on the  most widely RBF kernel SVM, with $ \gamma=0.001 $. We selected dog in the dataset for the positive classes and truck for the negative classes, a total of 4038.  We calculate the change of the fooling rate of adversarial perturbations for data with the change of $ \varepsilon $. Fig.~\ref{fig8} shows our results.
In the case of RBF kernel, the fooling rate increases with the increase of $\varepsilon $.
When $ \varepsilon $ is greater than $ 10^{-5} $, the fooling rate can reach $ 100\% $.
In the following experiments, we choose $ \varepsilon = 10^{-5} $.

\begin{figure}[h]  	\centerline{\includegraphics[width=0.35\textwidth]{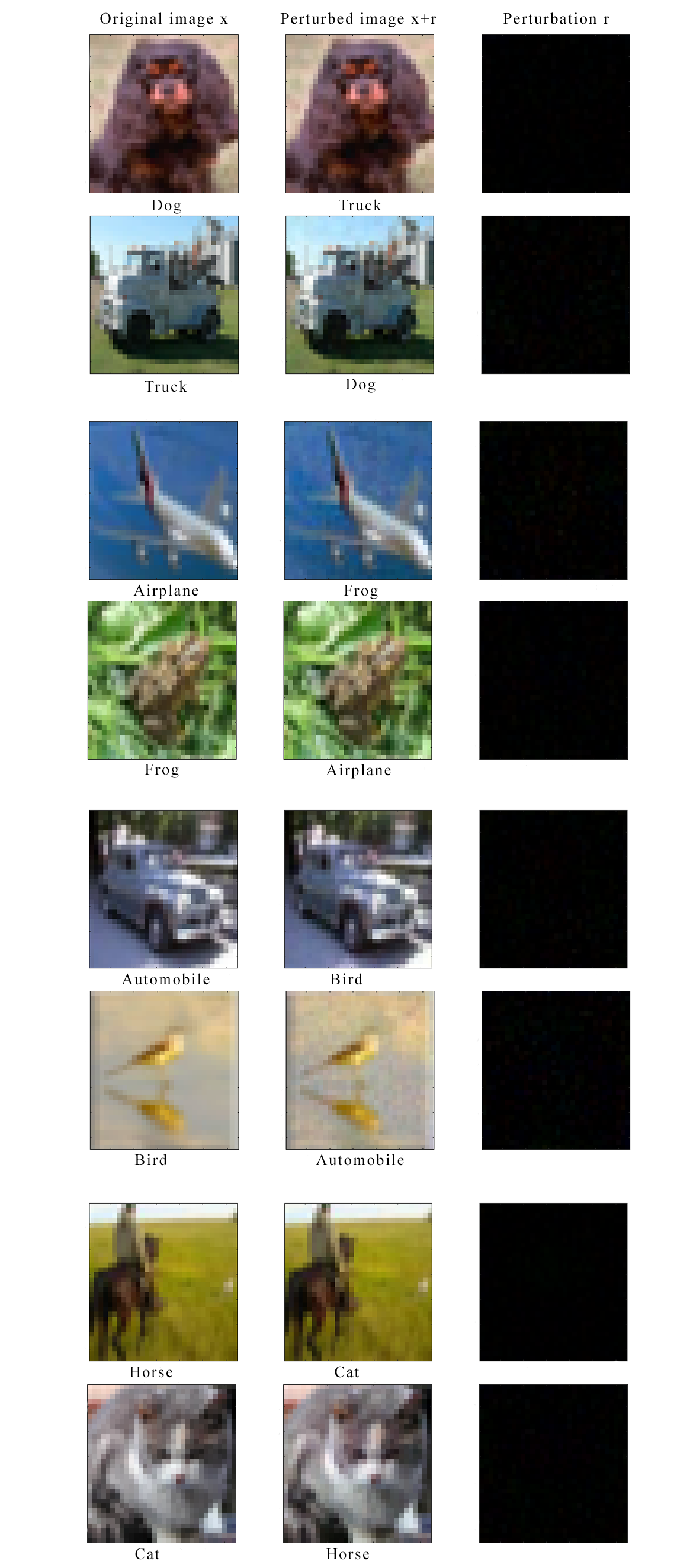}}  
\caption{The original images, the images that have been misclassified after being attacked, and the  images of sAP against RBF kernel SVM.}  
\label{fig9}  
\end{figure}

In  Fig.~\ref{fig9}, we select dog and truck, airplane and frog, automobile and bird, horse and cat as positive and negative classes pairs respectively, to build RBF kernel SVMs, and then solve the corresponding adversarial perturbations respectively. As shown in  Fig.~\ref{fig9}, the  first column are the original images, the second are the images after adding the attacks, and the third are the adversarial perturbations. We marked the classes of each image directly below the images. Through comparison, we find that the images after adding the attacks are basically the same as the original images in the human eyes, but the classes of the images after adding the attacks are actually opposite to the original images.

\section{Conclusions}\label{conclusion}

In this paper, we study the problem of adversarial perturbations for kernel SVMs.  In the past papers, the research on adversarial perturbation of machine learning models (especially nonlinear models) is basically generated based on a large number of iterative processes or approximate methods. These methods not only have a long iterative process, but also the attack is imprecise. At the same time, the researches on  adversarial perturbations  have been focused on simple generation, and it is difficult to find the essential relationship between attack and model. In this paper, we transform the original nonlinear optimization problem  into a nonlinear KKT system, which avoids the difficulties of the corresponding optimization problem and obtains effective and accurate results. In the numerical experiment part, we give the effective results of linear kernel, RBF kernel and polynomial kernel SVMs respectively. At the same time, it is observed from our visual experiment that the changes of parameters of RBF kernel and polynomial kernel will produce attacks with great differences.
In summary, on the one hand, accurate solution makes our method superior to other  advanced methods in calculating  adversarial perturbations (the fooling rate of our method can reach $ 100\% $). Further, accurate attack can avoid our operation from being found to the greatest extent, because it is the smallest attack vector compared with other methods. At this time, the small attack can be applied to the environment with practical significance when human cannot observe.
On the other hand, we also provide some insights into the influence of the parameters in the SVMs on the  adversarial perturbation, such as the relationship between the generated attacks and the linearity of the models, and the relationship between the parameters in the nonlinear models. In addition, based on our method, more research on improving the parameters of the model to enhance the robustness of the model to attacks can be carried out in the future. 
%Due to the implicit form of the nonlinear functions mapping data to the feature space, it is difficult to obtain the explicit form of the adversarial perturbations. Therefore, we transform the original optimization problem of attacking nonlinear SVMs into a nonlinear KKT system. In the numerical experiment part, we give the specific results of linear kernel, RBF kernel and polynomial kernel SVMs respectively. On one hand, we get that our method is practical and effective. On the other hand, we also provide some insights into the impact of parameters in SVMs on adversarial perturbations.

\vspace{12pt}

\end{document}